\def\year{2021}\relax
\documentclass[letterpaper]{article} % DO NOT CHANGE THIS
\usepackage{aaai21}  % DO NOT CHANGE THIS
\usepackage{times}  % DO NOT CHANGE THIS
\usepackage{helvet} % DO NOT CHANGE THIS
\usepackage{courier}  % DO NOT CHANGE THIS
\usepackage[hyphens]{url}  % DO NOT CHANGE THIS
\usepackage{graphicx} % DO NOT CHANGE THIS
\usepackage{natbib}  % DO NOT CHANGE THIS AND DO NOT ADD ANY OPTIONS TO IT
\usepackage{caption} % DO NOT CHANGE THIS AND DO NOT ADD ANY OPTIONS TO IT
\usepackage{algorithm}
\usepackage{amsmath,amssymb,amsfonts}
\usepackage{algpseudocode}
\floatname{algorithm}{Algorithm}

\usepackage{subfigure}
\usepackage[switch]{lineno}

\title{Dynamic Horizon Value Estimation for Model-based Reinforcement Learning}
\author{Junjie~Wang,\textsuperscript{\rm 1, \rm 2}
        Qichao~Zhang,\textsuperscript{\rm 1, \rm 2}
        Dongbin~Zhao,\textsuperscript{\rm 1, \rm 2}
        Mengchen~Zhao,\textsuperscript{\rm 3}
        Jianye~Hao \textsuperscript{\rm 3} \\
}

\affiliations {
    \textsuperscript{\rm 1} The State Key Laboratory of Management and Control for Complex Systems, Institute of Automation, \\Chinese Academy of Sciences, Beijing, China \\
    \textsuperscript{\rm 2} University of Chinese Academy of Sciences, Beijing, China \\
    \textsuperscript{\rm 3} Huawei Noah's Ark Lab, Beijing, China \\
}
\begin{document}
% \linenumbers

\maketitle

\begin{abstract}
Existing model-based value expansion methods typically leverage a world model for value estimation with a fixed rollout horizon to assist policy learning. However, the fixed rollout with an inaccurate model has a potential to harm the learning process. In this paper, we investigate the idea of using the model knowledge for value expansion adaptively. We propose a novel method called Dynamic-horizon Model-based Value Expansion (DMVE) to adjust the world model usage with different rollout horizons. Inspired by reconstruction-based techniques that can be applied for visual data novelty detection, we utilize a world model with a reconstruction module for image feature extraction, in order to acquire more precise value estimation. The raw and the reconstructed images are both used to determine the appropriate horizon for adaptive value expansion. On several benchmark visual control tasks, experimental results show that DMVE outperforms all baselines in sample efficiency and final performance, indicating that DMVE can achieve more effective and accurate value estimation than state-of-the-art model-based methods. 
\end{abstract}

\section{Introduction}
\label{sec:Intro}
Reinforcement learning (RL) has recently attracted wide attention due to its successful application in various fields with decision behavior, such as games \citep{shao2018learning,shao2018starcraft,tang2020starcraft}, robots navigation \citep{li2019deep}, and autonomous driving \citep{li2019reinforcement}. In general, RL methods can be divided into two categories: model-free RL methods and model-based RL methods. 
% Model-free RL approaches learn value function or policy directly without a world model, while model-based RL algorithms typically build a dynamics model that the agent uses to reason about decision behaviors. 
%  In Deep Reinforcement Learning (DRL), a neural network with reinforcement learning is used to enhance the algorithm the ability to control the system with extremely high-dimensional input spaces such as images (). Learning from limited samples is one of the challenges which can be faced when DRL is applied to a real-world System. Almost all real-world systems are either slow, fragile, or expensive enough that the data they produce is costly, and policy learning must be data-efficient (). Model-based reinforcement learning approaches make it possible to solve complex tasks given just a few training samples. 
For model-based RL methods, a dynamics model is usually bulit to help the learning process of an agent in two directions: direct policy learning \citep{sutton1991dyna} and value expansion \citep{feinberg2018model}. For the first direction, agents can interact with the learned model rather than the real environment to generate more experiences, so as to use less real data for policy learning.
% Such methods are often in the manner of Dyna \citep{sutton1991dyna}. The Dyna-style planning does not distinguish the real and simulated experiences when updating the value function and policy \citep{zaheer2020selective}. 
For the second direction, agents also interact with the built world model to predict states in the future sequence. The length of such sequence is called the rollout horizon. 
The difference is that, in this way, these imaginary data are not stored in the replay buffer to facilitate policy learning directly, but assist the value estimation. For the both two ways, a relatively accurate model is required. 
As is discovered in prior studies \citep{venkatraman2015improving,talvitie2017self}, even small model error can degrade multi-step rollouts seriously.
% due to the compounding of model error
% , and the policy optimization tends to exploit regions where the model is inaccurate
% According to \citet{feinberg2018model}, the model error can directly lead to the value estimation error. 
Thus, how to achieve better performance with an inaccurate world model becomes an important topic. 
In this paper, we focus on the second direction of using world models, which is the value expansion under an inaccurate model. 

\begin{figure}[t]
\centerline{\includegraphics[width=3.2in]{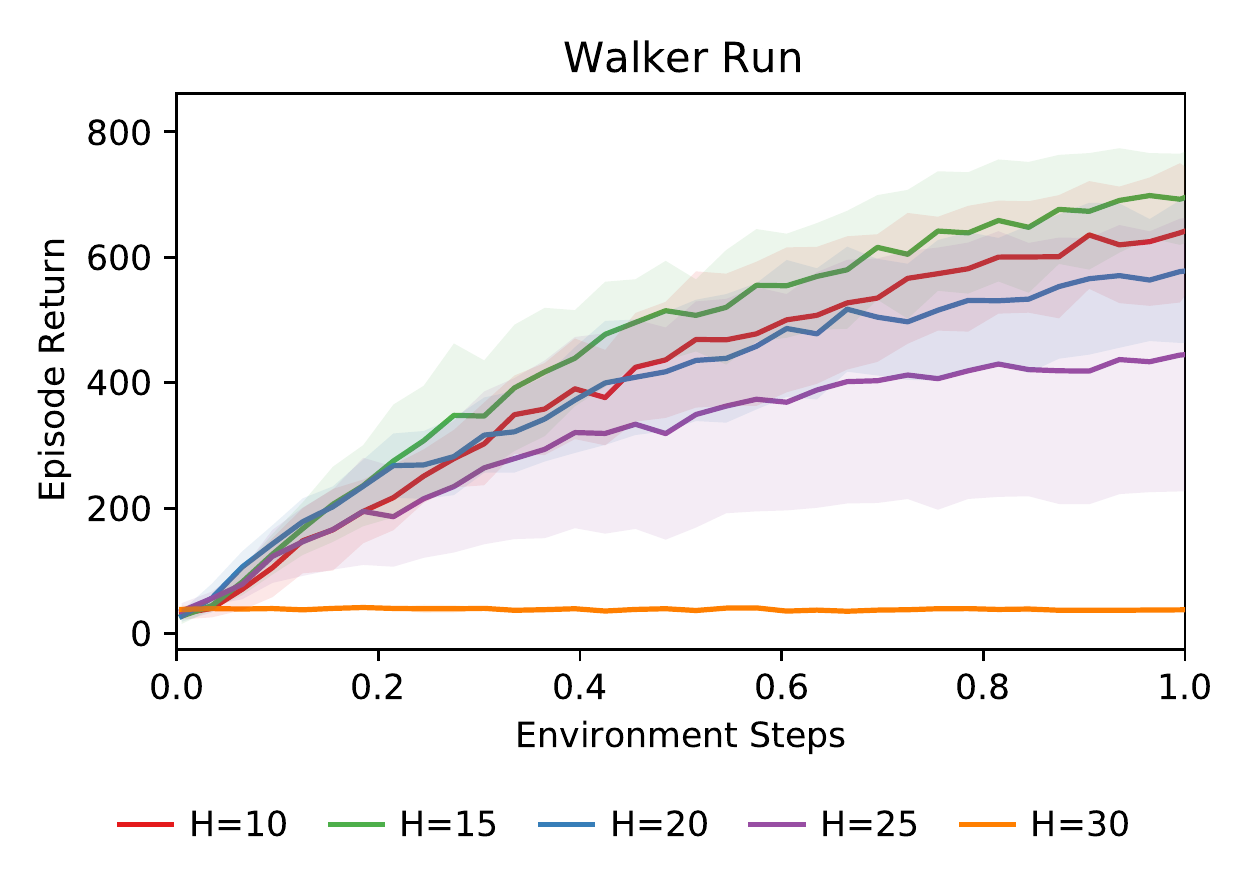}}
\caption{The learning curves of Dreamer \citep{hafner2019dream} with different rollout horizons $H$ on the task of Walker Run.  The solid lines depict the mean scores and shaded areas indicate the standard deviation of five trials over different random seeds. 
Each trial is evaluated every 10,000 environment steps. 
When applying relatively large $H$ (e.g.~$H=30$), Dreamer fails to learn a good policy on this task. Results indicate that the performance of Dreamer is sensitive to the setting of $H$. }
\label{DiffH}
\end{figure}

% To some extent, the uncertainty of a model reflects its accuracy. 
% Many of the existing model-based RL methods hope to obtain more accurate value estimation by incorporating multi-horizon rollouts with a fixed maximum one \citep{buckman2018sample,hafner2019dream,zaheer2020selective}. A demonstration of the relationship between the maximum rollout horizon selection and policy learning is illustrated in Figure~\ref{DiffH}. We evaluate the performance of Dreamer \citep{hafner2019dream} on a classic task of DeepMind Control Suite \citep{tassa2020dmcontrol} Walker Run. It can be seen that different rollout horizons have a great impact on the performance of the algorithm. Therefore, in this paper, we argue that more accurate value estimation can be gained by dynamically adjusting the rollout horizon of the model, and a state-related adaptive horizon should be given within the maximum.

Many of the existing model-based RL methods hope to obtain more accurate value estimation by incorporating multi-horizon rollouts, but the rollout length is fixed to the maximum setting. For example, STEVE \citep{buckman2018sample} leverages model ensembles to dynamically interpolating between different rollouts. Selective MVE \citep{zaheer2020selective} incorporates learned variance into planning aside from model ensembles. However, none of these works focuses on visual control tasks. When dealing with pixel input problems, approaches like STEVE may be computationally expensive. Dreamer \citep{hafner2019dream} focuses on visual control tasks. It also incorporates multi-horizon rollouts with fixed length by utilizing latent imagination. A demonstration of the relationship between the maximum rollout horizon selection and policy learning is illustrated in Figure~\ref{DiffH}. We evaluate the performance of Dreamer \citep{hafner2019dream} on a classic task of DeepMind Control Suite \citep{tassa2020dmcontrol} Walker Run. It can be seen that different rollout horizons have a great impact on the performance of the algorithm. Results indicate that while multi-step information can be integrated, the rollout horizon itself is still essential for model-based planning. Therefore, in this paper, we argue that more accurate value estimation can be gained by dynamically adjusting the rollout horizon of the model, and a state-related adaptive horizon should be given within the maximum. 

To deal with the influence of model inaccuracies, one natural solution is to incorporate uncertainty into the prediction of the world model.
Although model ensembles can be used to measure model uncertainty \citep{DBLP:conf/iclr/RajeswaranGRL17,kurutach2018model,clavera2018model}, they are not very suitable for visual control tasks due to the large computational burden. 
Intuitively, the uncertainty of the model and the novelty of data is highly relevant. In deep learning methods, when the new images are fed into the encoder-decoder models, these models tend to be ineffective and have significant reconstruction errors. Thus, the reconstruction error is considered as a measure of visual data novelty \citep{pimentel2014review,sabokrou2018adversarially,denouden2018improving}. Inspired by novelty detection with reconstruct-based approaches, the reconstructed images are used to determine the adaptive horizons for more accurate value estimation in our method. 

% To deal with the influence of model inaccuracies, one natural solution is to incorporate uncertainty into the prediction of the world model. 
% % Many of the existing model-based RL methods consider the model uncertainty in order to learn better decision behavior. 
% For example, model ensembles are often used to measure model uncertainty \citep{DBLP:conf/iclr/RajeswaranGRL17,kurutach2018model,clavera2018model}. 
% Intuitively, the uncertainty of the model and the novelty of data is highly relevant. For example, if a state never encountered in the model learning process, it is likely to be difficult for the model to cover this case, and its future imaginations for this novel state may be unreliable.
% Therefore, how to make good use of a model is the key to model-based RL. In deep learning, reconstruction models often rely on encoder-decoder deep neural networks to rebuild the input data. These models tend to be ineffective when encoding novel images and have significant reconstruction errors. Thus, the reconstruction error is considered as a measure of visual data novelty \citep{pimentel2014review,sabokrou2018adversarially,denouden2018improving}. Inspired by novelty detection with reconstruct-based approaches, the reconstructed images are used to determine the adaptive horizons for more accurate value estimation in our method. 
% % The significant error during the image reconstruction make the networks ineffective at compressing novel images. 

In this paper, we propose Dynamic-horizon Model-based Value Expansion (DMVE), which adjusts the world model usage for value estimation through adaptive rollout horizons selection. 
% Our starting point is very direct. Since the accuracy of the model is related to the uncertainty of the data, we can dynamically adjust the use degree of the model according to the size of the uncertainty of the data, that is, we can imagine the number of steps in the future. 
In order to estimate state values, DMVE first adopts a reconstruction network to rebuild the original observation. Subsequently, the raw and reconstructed images are both fed into the transition module to imagine in the latent space, and the value expansion is calculated with diverse imagination horizons for both. Afterward, the horizons corresponding to the top minimum value expansion errors between the original image and the reconstructed image are selected, and the final value estimation is averaged from the expansion values corresponding to the selected horizons. 
% The dynamic scheme of the world model in DMVE is reflected in two aspects: 1. As the training progresses, the accuracy of the world model and the novelty of the data changes dynamically, and so does the average value of the horizons selected by DMVE; 2. At any moment in the training process, the novelty of different input data is different, which leads to different accuracy of the model's future prediction of different states. 
Experimental results demonstrate that DMVE achieves higher sample efficiency and better asymptotic performance than baseline methods \citep{mnih2016asynchronous,barth2018distributed,hafner2019dream,hafner2019learning}. 

The contributions of our work are summarized as follows: 
\begin{itemize}
\item We develop an algorithm introducing Latent Imagination into Model-based  Value Expansion (MVE-LI) and leverages advanced value estimation and policy learning objectives for visual control tasks. 
\item We propose a novel method named DMVE, which can dynamically adjust the world model usage through adaptive rollout horizons selection. 
Furthermore, various experiments are designed to better understand the overall performance gained by the proposed algorithm. 
% \item We conduct experiments to demonstrate that our method outperforms several state-of-the-art approaches on benchmark visual control tasks. Furthermore, various experiments are designed to better understand the overall performance gained by our algorithm. 
\end{itemize}

\section{Related Work}
To solve reinforcement learning problems, the Policy Gradient (PG) methods are aimed at modeling and optimizing the policy directly. REINFORCE \citep{williams1992simple} is a classical PG method that relies on an estimated return by using episode samples to update the policy. In order to reduce gradient variance in vanilla policy gradients, many methods adopt actor-critic model to learn a value function in addition to the policy. For example, the critics in A3C \citep{mnih2016asynchronous} learn the value function while multiple actors are trained in parallel. 
% PPO (), a simplified version of TRPO (), uses a clipped surrogate objective and retains similar performance. SAC () incorporates the entropy measure of the policy into the reward to encourage exploration.
DDPG \citep{lillicrap2015continuous} combines DPG \citep{silver2014deterministic} with deep neural networks to learn a deterministic policy by experience replay. D4PG \citep{barth2018distributed}, an extension of DDPG, utilizes distributed parallel actors, distributional critic, multi-step returns, and prioritized experience replay. TD3 \citep{fujimoto2018addressing} is another variant of DDPG that applies clipped double Q-learning, delayed update, and target policy smoothing. 
SAC \citep{haarnoja2018soft} incorporates the entropy measure of the policy into the reward to encourage exploration. However, above methods fail to take advantage of gradients through transitions and simply maximize immediate values. In this work, our algorithm DMVE also employs the actor-critic model but it is in a model-based manner. 

Model-based reinforcement learning approaches intend to improve sample efficiency by learning a dynamics model to simulate the environment \citep{sutton2018reinforcement}. For instance, VPN \citep{oh2017value}, MVE \citep{feinberg2018model}, and Dreamer \citep{hafner2019dream} use the imagination of a learned model to assist the target value estimation. Dyna \citep{sutton1991dyna} and I2A \citep{racaniere2017imagination} learn dynamics to provide supplementary context for policy learning. Nevertheless, the model error has an inclination to harm the planning of model-based approaches, which is also known as model-bias \citep{deisenroth2011pilco}. This work is mainly based on the framework of Dreamer but we take the world model usage into account. 

Previous works also explore the model usage to deal with model-bias by considering the uncertainty of the learned model. 
% There are many possibilities to capture model uncertainty for model-based RL.
% Gaussian processes () and Bayesian neural networks incorporate uncertainty directly but are not suitable for complex tasks. Other Methods can be used to approximate model uncertainty like model ensembles () and dropout (). 
STEVE \citep{buckman2018sample} extends MVE by dynamically interpolating between model rollouts of various horizon lengths for each individual example. MBPO \citep{janner2019trust} as well as BMPO \citep{lai2020bidirectional} combines model ensembles with short model rollouts for sufficient policy optimization. Selective MVE \citep{zaheer2020selective} incorporates learned variance into planning to use the model selectively. 
There are several world model architecture alternatives, such as linear models \citep{parr2008analysis,sutton2012dyna,levine2014learning,kumar2016optimal}, Gaussian processes \citep{kuss2004gaussian,ko2007gaussian}, and neural networks \citep{draeger1995model,nagabandi2018neural}. World Models \citep{ha2018recurrent} learn latent dynamics to help evolve linear controllers. PlaNet \citep{hafner2019learning} learns world model components jointly and uses them to planning on latent space. In this work, DMVE utilizes the world model architecture in PlaNet \citep{hafner2019learning}. 

% TODO: PETS

\begin{figure*}[t]
\centerline{\includegraphics[width=6.8in]{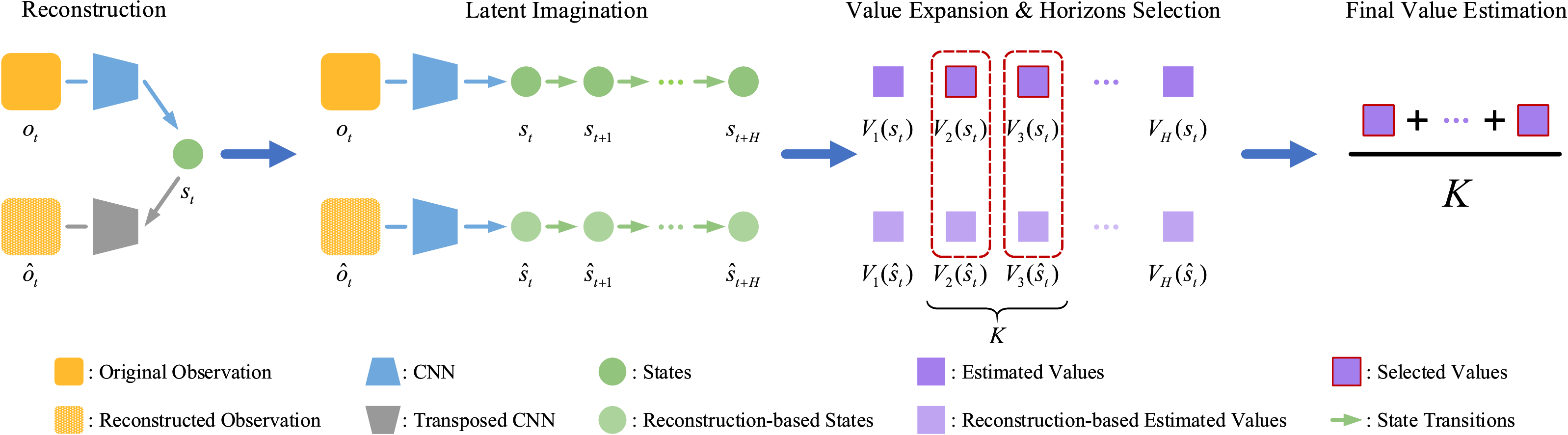}}
\caption{The overview of DMVE. To estimate state values, DMVE first adopts a reconstruction network to reconstruct the original observation. Then the raw and reconstructed images are both fed into the transition module to do $H$-step latent imagination, and the value expansion is calculated with diverse rollout horizons $h=1,2,\cdots, H$ for both. Afterward, the horizons corresponding to the top $K$ minimum value expansion errors between the original image and the reconstructed image are selected. Finally, the value estimation is averaged from the expansion values corresponding to the selected horizons. }
\label{Algo}
\end{figure*}

\section{Preliminaries}
\subsection{POMDP and RL}
In this paper, the task of visual control is formulated as a Partially Observable Markov Decision Process (POMDP) since the agent cannot directly observe the underlying states on this kind of task. POMDP is a generalization of Markov Decision Process (MDP), which connects unobservant system states to observations. Formally, it can  be described as a 7-tuple $\left( \mathcal{S}, \mathcal{A}, \mathcal{O}, T, R, P, \gamma \right)$, where $\mathcal{S}$ denotes the set of states, $\mathcal{A}$ denotes the set of actions, and $\mathcal{O}$ denotes the set of observations. The agent interacts with the environment at each of a sequence of discrete time steps, $t=0,1,\cdots$. $T \left(s_{t+1}|s_{t}, a_{t} \right)$ is the conditional transition probability that action $a_{t} \in  \mathcal{A}$ in state $s_{t} \in \mathcal{S}$ will lead to state $s_{t+1}$. $R(s_t, a_t)$ yields the real-valued reward for executing action $a_t$ in state $s_t$ and $P$ denotes the observation probabilities $P(o_{t}|s_{t+1}, a_{t})$, where $o_t$ stands for the observation made by the agent, when action $a_{t}$ is executed and the world moved to the state $s_{t+1}$. \(\gamma \in (0,1)\) is a discount factor. 

The policy $\pi(s)$ maps from environmental states to actions. At each time step $t$, the environment is in some state \(s_t \in \mathcal{S}\), and the agent selects a feasible action \(a_t \in \mathcal{A}\) in one state, which causes the environment to transition to state \(s_{t+1} \in \mathcal{S}\) with probability $T \left(s_{t+1}|s_{t}, a_{t} \right)$. After the action carried out in the environment, the agent receives an observation $o_t \in \mathcal{O}$ with probability $P(o_{t}|s_{t+1}, a_{t})$ and a numerical reward $R(s_t, a_t)$. Then the above interaction process repeats.
The goal of RL is to learn an optimal policy by maximizing the cumulative reward \citep{sutton2018reinforcement}, 
% that is, 
\begin{equation}
\max \mathbb{E}\left[ {\sum\limits_{t = 0}^\infty  {{\gamma ^t}R\left( {{s_t},{a_t}} \right)} } \right]. 
\end{equation}

\subsection{MVE}
\label{sec:MVE}
Model-based Value Expansion \citep[MVE;][]{feinberg2018model} utilizes a learned dynamics model to enhance value estimates, hoping to improve both the sample efficiency and performance. 

Using the imagined states $\{s_{t+n}\}_{n = 1}^{H}$ (where $H$ denotes the finite maximum rollout horizon) obtained from the world model, the $H$-step value expansion of a given state $s_t$ can be estimated with
\begin{equation}
{V_H}({s_t}) = \sum\limits_{n = 0}^{H - 1} {{\gamma ^n}{r_{t + n}}}  + {\gamma ^H}v({s_{t + H}}),
\end{equation}
where $v(s)$ denotes the value function and $r$ denotes a state-related reward function. 

\begin{algorithm*}[ht]
  \caption{Dynamic-horizon Model-based Value Expansion (DMVE)}
  \begin{algorithmic}[1]
    \State Initialization: Number of random seed episodes $S$, collect interval $C$, batch size $B$, sequence length $L$, maximum imagination horizon $H$, learning rate $\alpha$, and number of selected horizons $K$
    \State Collect dataset $\mathcal{D}$ with $S$ episodes through the interaction with the environment ENV using random actions
    \State Initialize world model $w_\theta$, actor network $\pi_\phi$, critic network $v_\psi$
    \While{not converged}
      \For{$c = 1,\cdots, C$}
        \State Sample $B$ data sequences \(\left\{ {\left( {{a_t},{o_t},{r_t}} \right)} \right\}_{t = k}^{k + L} \sim \mathcal{D}\) and derive model states \(s_t \sim w_\theta ^S\left(s_t|s_{t-1}, a_{t-1}, o_t\right)\)
        % \State Derive model states \(s_t \sim w_\theta ^S\left(s_t|s_{t-1}, a_{t-1}, o_t\right)\)
        \State Update $\theta$ using representation learning
        \State Compute the reconstructed observations \({\hat o_t} \sim w_\theta ^O\left( {{\hat o_t}|{s_t}} \right)\) and reconstruction-based states \(\hat s_t \sim w_\theta ^S\left(\hat s_t|s_{t-1}, a_{t-1}, \hat o_t\right)\)
        \State Estimate state values $V_h({s_t})$ and $V_h({\hat s_t})$ following Eq.~\eqref{ValueEstimation} for $h=1,\cdots,H$
        \State Select $K$ horizons via Eq.~\eqref{SelectH} and determine final value estimation $V({s_t})$ via Eq.~\eqref{FinalV}
        % \State Update actor network parameters \(\phi  \leftarrow \phi  + \alpha {\nabla _\phi }\left( {\max {{\left\{ {{V_h}({s_t})} \right\}}_{h = 1, 2, \cdots,H}}} \right)\)
        % \State Update critic network parameters \(\psi  \leftarrow \psi  - \alpha {\nabla _\psi }\left( {\frac{1}{2}{{\left\| {{v_\psi }({s_t}) - {V}({s_t})} \right\|}^2}} \right)\)
        \State Update actor network parameters \(\phi  \leftarrow \phi  + \alpha {\hat \nabla _\phi }J_\pi \left(\phi \right)\)
        \State Update critic network parameters \(\psi  \leftarrow \psi  - \alpha {\hat \nabla _\psi }J_v \left(\psi \right)\)
      \EndFor
      \State \({o_1} \leftarrow \text{ENV.reset}\left(\right)\)
      \For{$t = 1,\cdots, T$}
        \State Compute \(s_t \sim w_\theta ^S\left(s_t|s_{t-1}, a_{t-1}, o_t\right)\) from history and \(a_t \sim \pi_\phi(a_t|s_t)\) with the actor network
        \State Add exploration noise to action and take it in environment to get  \({r_t,o_{t+1}} \leftarrow \text{ENV.step}\left(a_t\right)\)
      \EndFor
      \State Add experience to dataset \(\mathcal{D} \leftarrow \mathcal{D} \cup \left\{ {\left( {{a_t},{o_t},{r_t}} \right)} \right\}_{t = 1}^T\)
    \EndWhile
  \end{algorithmic}
  \label{DMVE}
\end{algorithm*}

\section{Methodology}
% In this section, we present the framework of DHVE and further introduce how DHVE dynamically adjusts the world model usage through the uncertainty evaluation of data in visual control tasks. Despite dynamic-horizon value estimation can be integrated into many model-based RL approaches, we choose Dreamer [] as the framework backbone since it learns long-horizon behaviors from images and leverages neural network dynamics for value estimation.
In this section, we present the framework of DMVE and further introduce how DMVE dynamically adjusts the world model usage. The overall architecture is illustrated in Figure~\ref{Algo}, and the pseudocode is shown in Algorithm~\ref{DMVE}. 

\subsection{MVE by Latent Imagination}
We first present our framework backbone, an MVE-like algorithm that takes advantage of a world model, an action model, and a value model to estimate state values. MVE is proposed by \citet{feinberg2018model} which builds a dynamics model and uses its imagination as the context for value estimation (see Section~\ref{sec:MVE}). Here, we extend MVE to visual control tasks by adopting latent imagination \citep{hafner2019dream}, and refer to this algorithm as MVE-LI (where LI stands for Latent Imagination) in this paper. 

\subsubsection{World Model}
Many model-based RL methods first build a world model and further use it to derive behaviors. In the case that the world model is learned, the process of model learning and policy learning can be alternately paralleled. Usually, the world model provides the dynamics of a system, mapping from current state and action to the next state and giving rewards for this transition. There are several world model alternatives having the ability to be applied to visual control tasks, we choose the one used in PlaNet \citep{hafner2019learning} and Dreamer \citep{hafner2019dream} since it learns dynamics for planning by reconstructing the original images. At the same time, this reconstruction-based architecture is suitable for dynamic horizons selection (see Section~\ref{sec:DMVE}). The world model $w_\theta$ is made up of the following components, 
\begin{align}
\begin{aligned}
&\text{Representation module:}&&s_t \sim w_\theta ^S \left(s_t|s_{t-1}, a_{t-1}, o_t\right)\\
&\text{Reconstruction module:}&&\hat o_t \sim w_\theta ^O \left(\hat o_t | s_t \right)\\
% &\text{Reconstruction module:}&&o_t \sim w_\theta ^O \left(o_t | s_t \right)\\
&\text{Reward module:}&&r_t \sim w_\theta ^R \left(r_t | s_t \right)\\
&\text{Transition module:}&&s_t \sim w_\theta ^T \left(s_t|s_{t-1}, a_{t-1} \right), 
\end{aligned}
\end{align}
where $\hat o_t$ denotes the reconstructed observation of time step $t$. 
% \begin{equation}
% &\text{Representation module:} & w_\theta ^s() \\
% &Observation module: \\
% &Reward module: \\
% &Transition module: 
% \end{equation}

In the POMDP setting, states cannot be obtained directly, the representation module is thus applied to map the observations with actions to low-dimensional continuous vectors that are regarded as the states of Markovian transitions \citep{watter2015embed,zhang2019solar,hafner2019dream,hafner2019learning}. The reconstruction module estimates the original observations from the states and ensures that the states can represent effective information in the raw input data by minimizing the reconstruction error. The reward module predicts the rewards in the imaginary trajectories based on the real-value rewards from the environmental feedback. And the transition module predicts the next state purely depending on the current state and action without seeing the raw observation. 
In this work, the transition module is implemented as a Recurrent State Space Model  \citep[RSSM;][]{hafner2019learning}. The representation module is a combination of the RSSM and a Convolutional Neural Network \citep[CNN;][]{lecun1989backpropagation}. The reconstruction module is a transposed CNN, and the reward module is a dense network. 
% The overall parameter vector $\theta$ is updated through stochastic backpropagation [].
All four model components are optimized jointly through stochastic backpropagation, which is the same as Dreamer \citep{hafner2019dream}. 
% Please refer to [] for more details. 

% \textbf{Action and value Models}
% Consider imagined trajectories with a finite horizon H . Dreamer uses an actor critic approach to learn behaviors that consider rewards beyond the horizon. We learn an action model and a value model in the latent space of the world model for this. The action model implements the policy and aims to predict actions that solve the imagination environment. The value model estimates the expected imagined rewards that the action model achieves from each state s ,
% Action model:
% Value model:

% The action and value models are trained cooperatively as typical in policy iteration: the action model aims to maximize an estimate of the value, while the value model aims to match an estimate of the value that changes as the action model changes.
% We use dense neural networks for the action and value models with parameters  and  , respectively. The action model outputs a tanh-transformed Gaussian (Haarnoja et al., 2018) with sufficient statistics predicted by the neural network. This allows for reparameterized sampling (Kingma and Welling, 2013; Rezende et al., 2014) that views sampled actions as deterministically dependent on the neural network output, allowing us to backpropagate analytic gradients through the sampling operation,. 

\subsubsection{Policy Learning}
\label{sec:PolicyLearning}
We adopt an actor-critic method for policy learning. 
In addition to the estimated value function, actor-critic methods have an independent structure to represent the policy. The policy structure is known as the actor, since is used to derive behaviors. At the same time, the value function is called a critic because it criticizes the behaviors decided by the actor. 
% Actor-critic methods have a separate memory structure to explicitly represent the policy independent of the value function. The policy structure is known as the actor, since it is used to select actions. Meanwhile, the estimated value function is known as the critic, because it criticizes the actions made by the actor. Learning is always on-policy: the critic must learn about and critique whatever policy is currently being followed by the actor. 
For state $s_t$, the actor and critic models are defined as,
\begin{align}
\begin{aligned}
&\text{Actor:} &&a_t \sim \pi_\phi(a_t|s_t)\\
&\text{Critic:} &&v_\psi({s_t}) \approx {\mathbb{E}_{\pi ( \cdot |{s_t})}}\left( {\sum\limits_{\tau  = t}^{t + H} {{\gamma ^{\tau  - t}}{r_\tau }} } \right), 
\end{aligned}
\end{align}
In this work, the actor and critic models are both implemented as dense networks \citep{hafner2019dream}. 
State values need to be estimated for the actor and critic models optimization. MVE can improve value estimation by assuming an approximate dynamics model and a reward function. Since the aforementioned world model contains the elements needed for value expansion, we can use it to estimate state values. While several approaches can be adopted, MVE-LI uses the classic value expansion introduced in MVE paper \citep[see Section \ref{sec:MVE}]{feinberg2018model} . Selecting one $h \in \{1,2, \cdots, H\}$, we can compute the value estimation of $s_t$ with imagined trajectories $\left\{ {\left( {{s_{t+n}},{a_{t+n}},{r_{t+n}}} \right)} \right\}_{n = 0}^{H}$ using
\begin{align}
\begin{aligned}
V_h({s_t}) &= V(s_t,h)\\
&= {\mathbb{E}_{{w_\theta },{\pi _\phi }}}\left(\sum\limits_{n = 0}^{h - 1} {{\gamma ^n}{r_{t + n}} + {\gamma ^h}{v_\psi }({s_{t + h}})}\right), 
\end{aligned}
\end{align}
where the expectation is estimated under the imagination. 

In MVE-LI, the imagination horizon is fixed to the maximum $H$. Therefore, the estimated value for state $s_t$ can be represented as $V_H({s_t})$. After estimating the state values, the actor and critic neural networks can be optimized while the world model is fixed. The learning objectives of the actor and critic models in MVE-LI are set as, 
\begin{align}
{J_\pi }(\phi ) &= \mathop {\max }\limits_\phi  {\mathbb{E}_{{w_\theta },{\pi _\phi }}}\left( {\max {{\left\{ {{V_h}({s_t})} \right\}}_{h = 1, 2, \cdots,H}}} \right), \label{LearningObjA}\\
{J_v}(\psi ) &= \mathop {\min }\limits_\psi  {\mathbb{E}_{{w_\theta },{\pi _\phi }}}\left( {\frac{1}{2}{{\left\| {{v_\psi }({s_t}) - {V}({s_t})} \right\|}^2}} \right), 
\label{LearningObjV}
\end{align}
where the estimated value ${V}({s_t})={V_H}({s_t})$ for MVE-LI.

\subsection{Dynamic Horizon Value Estimation}
\label{sec:DMVE}
The developed MVE-LI imagines future states in the latent space with a fixed maximum rollout horizon $H$. However, for different imagination horizons $h \in \left\{1, 2, \cdots, H \right\}$, we can estimate diverse values for state $s_t$,
\begin{equation}
V_h({s_t}), h \in \left\{1, 2, \cdots, H \right\}.
\end{equation}
This leads to the question of which $h$ is the most suitable for the current state $s_t$.

\subsubsection{Basic Assumptions}
A reconstruction model often relies on encoder-decoder deep neural networks to rebuild the input data. These networks usually design a hidden layer with a lower dimensionality than the original input, and are optimized to learn a compressed representation of the input data distribution. Therefore, these models have a tendency to be ineffective at encoding novel data and fail to reconstruct the input without significant error \citep{pimentel2014review,sabokrou2018adversarially,denouden2018improving}. Thus, the reconstruction error is considered as a measure of visual data novelty. 
% A reconstruction model often relies on encoder-decoder deep neural networks to rebuild the input data. These networks usually design a hidden layer with a lower dimensionality than the original input and are optimized to learn a compressed representation of the input data distribution. Therefore, these models should be ineffective when encoding new images and cannot reconstruct the input without obvious errors. Therefore, reconstruction error is considered a measure of novelty.

Coincidentally, in a model-based RL setting, the data novelty has the potential to influence the value estimation accuracy since the world model may fail to predict trustable future states of novel data. This inspired us to apply the reconstruction-based approaches to dynamically adjust the world model usage with different rollout horizons for different states. Based on the observation of novelty detection with reconstruction error, the output difference of a model between the original image and the reconstructed image may also reflect the generalization ability of the model on the current input. To sum up, we enumerate the basic assumptions of DMVE here, 
\begin{itemize}
\item For MVE, the data novelty influences the value estimation accuracy; 
\item The output error between the original and the reconstructed images of the world model reflects the generalization of the model to the observation.
\end{itemize}

\begin{figure*}[t]
\centerline{\includegraphics[width=7.1in]{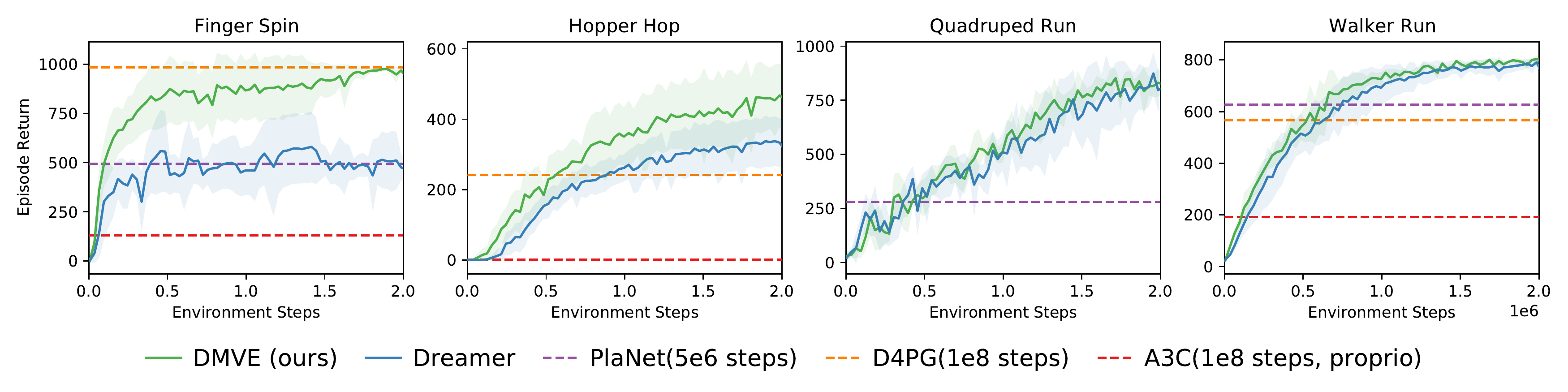}}
\caption{Comparison of DMVE (ours) and four baselines on different continuous visual control environments. The solid lines depict the mean scores and shaded areas indicate the standard deviation of five trials over different random seeds. Each trial is evaluated every 10,000 environment steps. The performance of D4PG, A3C, and PlaNet are taken directly from Dreamer paper.}
\label{curves}
\end{figure*}

\subsubsection{Reconstruction-based Horizon Selection}
Since we formulate the task of visual control as a POMDP problem, the original observations cannot fully characterize the states. Therefore, it is unreasonable to directly determine the imagination horizon with the reconstruction error. Based on the aforementioned assumptions, we now describe an algorithm using reconstructed images to do dynamic horizon value estimation. 

Given the maximum rollout horizon $H$, we can get the value estimation $V_h$ with different $h \in \left\{1, 2, \cdots, H \right\}$ for the data sequences \({ \mathop {\left\{ {\left( {a_t,o_t,r_t} \right)} \right\}}\nolimits_{t = k}^{k + L} }\) using the world model, where $L$ denotes the sequence length. First, the state of time step $t$ can be derived using $s_t \sim w_\theta ^S\left(s_t|s_{t-1}, a_{t-1}, o_t\right)$. Subsequently, the value of state $s_t$ can be estimated following
% \begin{equation}
% s_t \sim w_\theta ^S\left(s_t|s_{t-1}, a_{t-1}, o_t\right), 
% \end{equation}
\begin{align}
\begin{aligned}
%  \tau &= t,t+1,\cdots,t+H\\
% s_\tau &\sim w_\theta ^T\left(s_\tau|s_{\tau-1}, a_{\tau-1}\right)\\
% % r_\tau &= {\mathbb{E}_{w_\theta ^R\left( { \cdot |{s_\tau }} \right)}}\left( {{r_\tau }|{s_\tau }} \right), 
% r_\tau &\sim w_\theta ^R\left(r_\tau|s_\tau\right)\\
% v_\tau &\sim v_\psi \left(s_\tau\right), \tau = t,t+1,\cdots,t+H\\
&\underbrace {
\begin{aligned}
s_\tau &\sim w_\theta ^T\left(s_\tau|s_{\tau-1}, a_{\tau-1}\right)\\
r_\tau &\sim w_\theta ^R\left(r_\tau|s_\tau\right)\\
v_\tau &\sim v_\psi \left(s_\tau\right)
\end{aligned}
}_{\tau = t,t+1, \cdots ,t+H} \\
V_h(&{s_t}) = \sum\limits_{\tau = t}^{t+h - 1} {{\gamma ^{\tau-t}}{r_\tau} + {\gamma ^h}{v_\psi }({s_{t + h}})}. 
\label{ValueEstimation}
\end{aligned}
\end{align}
Similarly, for the reconstructed image $\hat o_t \sim w_\theta ^O\left(\hat o_t|s_t \right)$, the reconstruction-based state can be derived with $\hat s_t \sim w_\theta ^S\left(\hat s_t|s_{t-1}, a_{t-1}, \hat o_t\right)$
% \begin{equation}
% \hat s_t \sim w_\theta ^S\left(\hat s_t|s_{t-1}, a_{t-1}, \hat o_t\right)
% \end{equation}
and the corresponding value can be estimated by following the same process of Eq.~\eqref{ValueEstimation}. Now we have $V_h({s_t})$ as well as $V_h({\hat s_t})$ for $h=1,2,\cdots,H$. On top of that, given the number of selected horizons $K$, the set of rollout horizons $\mathcal{H}$ can be determined by 
\begin{align}
\begin{aligned}
\mathcal{V} &= { \left\{ {\left| {{V_h}\left( {{s_t}} \right) - {V_h}({{\hat s}_t})} \right|} \right\}_{h = 1}^{H}} \\
\mathcal{H} &= top_{\arg \min} \left( { \mathcal{V},K} \right), 
\label{SelectH}
\end{aligned}
\end{align}
where the $top_{\arg \min} \left( \mathcal{V},K \right)$ operation means to get the set of $K$ indices corresponding to the top $K$ minimum values in set $\mathcal{V}$. Then the final value estimation of DMVE is
\begin{equation}
V({s_t}) = \frac{1}{K}\sum\limits_{h \in \mathcal{H}} {{V_h}({s_t})}.
\label{FinalV}
\end{equation}
And now, the actor and critic models can be optimized following the learning objectives in \eqref{LearningObjA} and \eqref{LearningObjV}, respectively. 

In the above process of adaptive horizons selection, the original and the reconstructed images are fed into different world model components, to predict future sequence and perform value estimation, respectively. According to our assumptions, the output error of the world model between these images reflects its generalization ability of different inputs. As a result, the error between the estimated state values based on the two is used to determine the horizons of the final value estimation. For stability consideration, we set a hyperparameter $K$, which is the number of selected horizons. The evaluation of how this hyperparameter influences performance can be found in Section~\ref{sec:Eval}. 

\section{Experiments}
\label{sec:Exp}

Our experiments aim to study the following three primary questions: 1) How well does DMVE perform on benchmark reinforcement learning visual control tasks, comparing with state-of-the-art model-based and model-free methods?  2) What are the critical components of DMVE? 3) How do our design choices affect the performance?

\subsection{Comparative Evaluation}

In this section, we compare our method with prior baselines, and these 
methods represent the state-of-the-art in both model-free and model-based reinforcement learning. Specifically, for model-free methods, the baselines include A3C \citep{mnih2016asynchronous} and D4PG \citep{barth2018distributed}. For model-based approaches, we compare against PlaNet \citep{hafner2019learning}, which directly uses the world model for planning rather than explicit policy learning, and Dreamer \citep{hafner2019dream}, learns long-horizon behaviors from images and leverages neural network dynamics for value estimation, which is highly related to our approach. 

We evaluate DMVE and these baselines on 4 visual control tasks of the DeepMind Control Suite \citep{tassa2020dmcontrol}, namely, Finger, Hopper, Quadruped, and Walker. These tasks provide various challenges, for examlpe, the Hoppe Hop task delivers a difficulty in terms of sparse rewards. 
% These tasks pose a variety of challenges, including sparse rewards and contact dynamics. 
Observations are images of shape $64 \times 64 \times 3$. Each episode lasts for 1000 time steps with random initial states. 
% All our experiments are implemented on a computer with an Intel Xeon E5-2620 CPU and Nvidia GeForce RTX2080TI GPU. 

% In this section, all the hyperparameters listed in Algorithm~\ref{DMVE} including the neural network architectures are the same as the settings in the Dreamer paper \citep{hafner2019dream}, except for the number of selected horizons $K$ (appears in \eqref{SelectH} and is set to $3$), which is not available in Dreamer since its rollout horizon is fixed. 
In this section, the number of selected horizons $K$ is set to $3$. 
% More experimental details can be found in Supplementary Material. 
The comparison results are illustrated in Figure~\ref{curves}. Across diverse visual control tasks, our method DMVE learns faster than existing methods and achieves better asymptotic performance than previous model-based and model-free algorithms. Take the Finger Spin task as an example, DMVE exceeds Dreamer by a wide margin, and gains comparable performance at 2 million steps as that of D4PG at 100 million steps. On the Hopper Hop task that requires long-horizon credit assignment, DMVE surpasses all baselines as well. Our experiments indicate that DMVE can achieve more effective and accurate value estimation than state-of-the-art model-based method Dreamer\footnote{We rerun Dreamer on our device using the code publicly released by the authors at https://github.com/danijar/dreamer.}. 

\subsection{Design Evaluation}
\label{sec:Eval}
To better understand where the performance of DMVE benefits, different experiments are designed to evaluate our algorithm.

\begin{figure}[t]
\centerline{\includegraphics[width=3.3in]{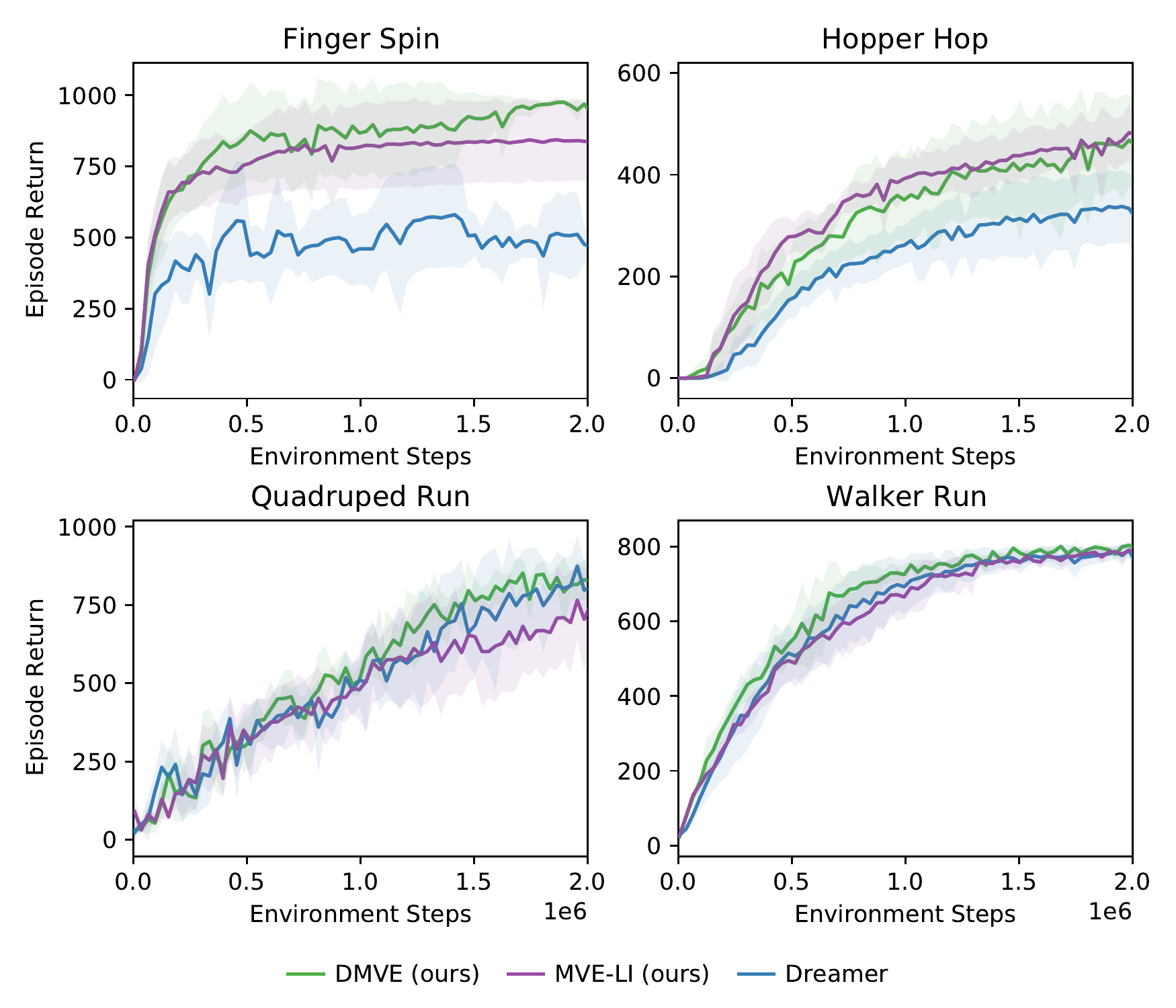}}
\caption{Comparison of DMVE and MVE-LI. The results are averaged from five trials. Each trial is evaluated every 10,000 environment steps. }
\label{compare}
\end{figure}

\subsubsection{Fixed Horizon}
In this section, on the aforementioned experimental tasks, we first compare DMVE with our framework backbone MVE-LI which utilizes the same learning objectives but estimates state values with a fixed rollout horizon, to understand the contribution of the policy learning part and the dynamic horizon selection part of DMVE. The results are shown in Figure~\ref{compare}. Surprisingly, MVE-LI also surpasses Dreamer by a large margin in the Finger Spin and Hopper Hop tasks. However, on the tasks of Quadruped Run and Walker Run, its performance is not as good as Dreamer. It can be seen from the experimental results that our basic method MVE-LI uses advanced value estimation and policy learning objectives to achieve good results, and DMVE further improves the performance on its basis. And results indicate that the overall performance of DMVE benefits from two parts, one is the value estimation and the optimization of actor and critic networks, and the other is the dynamic rollout horizon selection for value estimation. Contributed by these two parts, DMVE basically achieves the best performance on the various challenging tasks according to our experiments. Although the learning efficiency is slightly reduced after adding dynamic horizon value estimation to the Hopper Hop task, the approximate best performance is finally obtained.

% \subsubsection{Dynamic Horizon Analysis}
% Further, we show the curves of the actor loss and the average of the selected horizon during training. 

\subsubsection{Value Estimation}
In this work, we make use of a different value estimation from Dreamer  (see Section~\ref{sec:PolicyLearning}). To be specific, Dreamer uses $V_ \lambda$, that is, 
\begin{align}
\begin{aligned}
V_N^k({s_\tau }) &= {\mathbb{E}_{{w_\theta },{\pi _\phi }}}\left( {\sum\limits_{n = \tau }^{h - 1} {{\gamma ^{n - \tau }}{r_n} + } {\gamma ^{h - \tau }}{v_\psi }\left( {{s_h}} \right)} \right),\\
{V_\lambda }({s_t}) &= \left( {1 - \lambda } \right)\sum\limits_{n = 1}^{H - 1} {{\lambda ^{n - 1}}V_N^n({s_t}) + } {\lambda ^{H - 1}}V_N^H({s_t}), 
\end{aligned}
\end{align}
where $\tau  = t,t + 1, \cdots ,t + H, h = \min \left( {\tau  + k,t + H} \right)$. We do not find incorporating $V_ \lambda$ with DMVE works well because $V_ \lambda$ itself integrates multi-horizon values. This reflects to a certain extent that it is more advantageous to select a suitable horizon adaptively for value estimation. 

\subsubsection{Hyperparameters Study}
In this section, we study the question of how our design choices affect the performace of DMVE. We choose two key hyperparameters of our DMVE algorithm, namely, the maximum rollout horizon $H$, which is used in Eq.~\eqref{ValueEstimation}, and the number of selected horizons $K$, which appears in Eq.~\eqref{SelectH}. To be specific, the hyperparameters of different experimental runs here are set as, 
\begin{align}
&H=10,15,20,25,30, &&K=3,\\
&K=1,3,5,7,9, &&H=15.
\end{align}

\begin{figure}[ht]
\centerline{\includegraphics[width=3.0in]{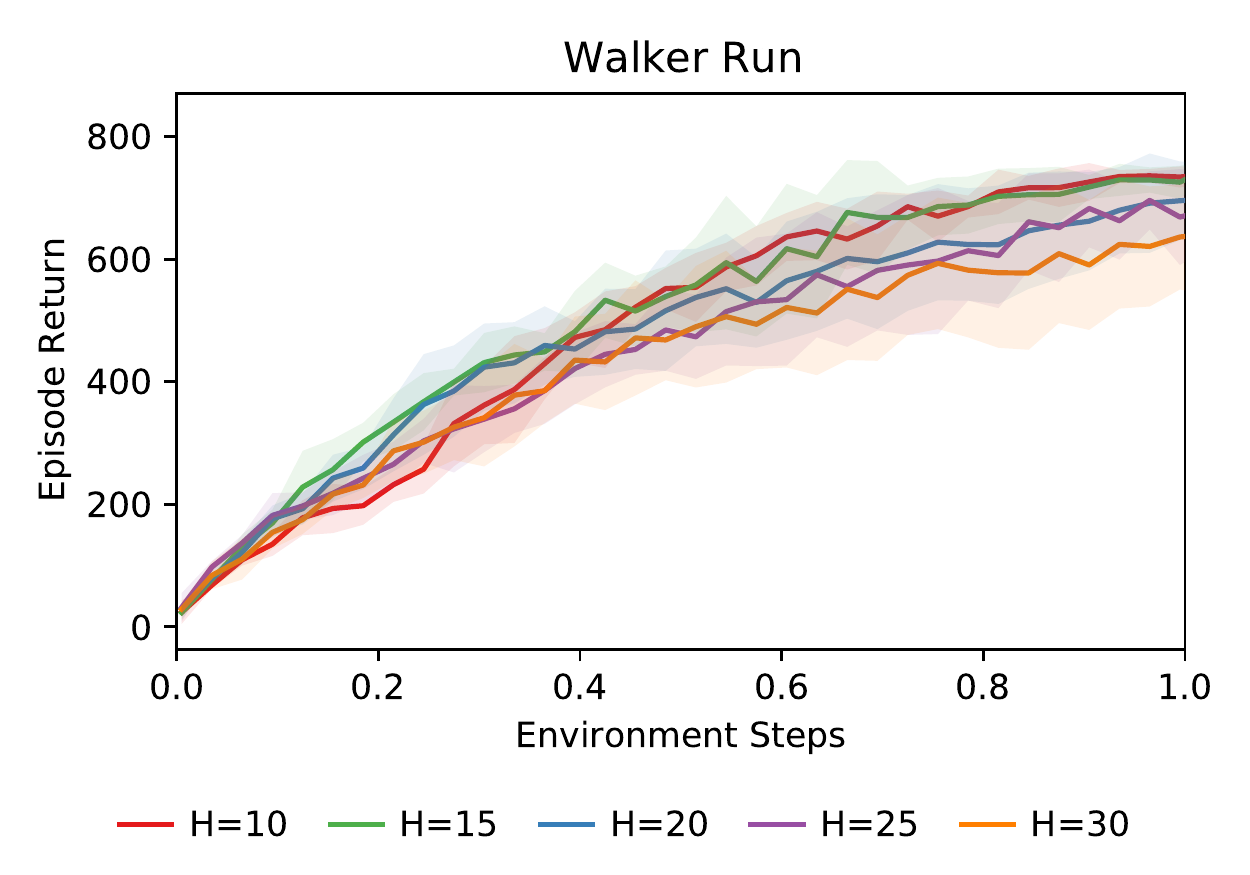}}
\caption{The learning curves of DMVE with different maximum imagination horizons $H$ on the task of Walker Run. The solid lines depict the mean scores and shaded areas indicate the standard deviation of five trials over different random seeds. Each trial is evaluated every 10,000 environment steps.  }
\label{DMVEH}
\end{figure}
First, we evaluate DMVE with different $H$ ($K$ is fixed to $3$ here across different runs), and the results are shown in Figure~\ref{DMVEH}. From the comparison between Figure~\ref{DMVEH} and Figure~\ref{DiffH} (illustrated in Section~\ref{sec:Intro}), it is obvious that DMVE has better performance retention for various $H$ than Dreamer. When applying with long-horizon imagination (e.g.~$H=30$), DMVE still can learn a relatively good policy on this task. This demonstrates the importance of adaptive horizon selection for value estimation, and policy learning. 

\begin{figure}[ht]
\centerline{\includegraphics[width=3.3in]{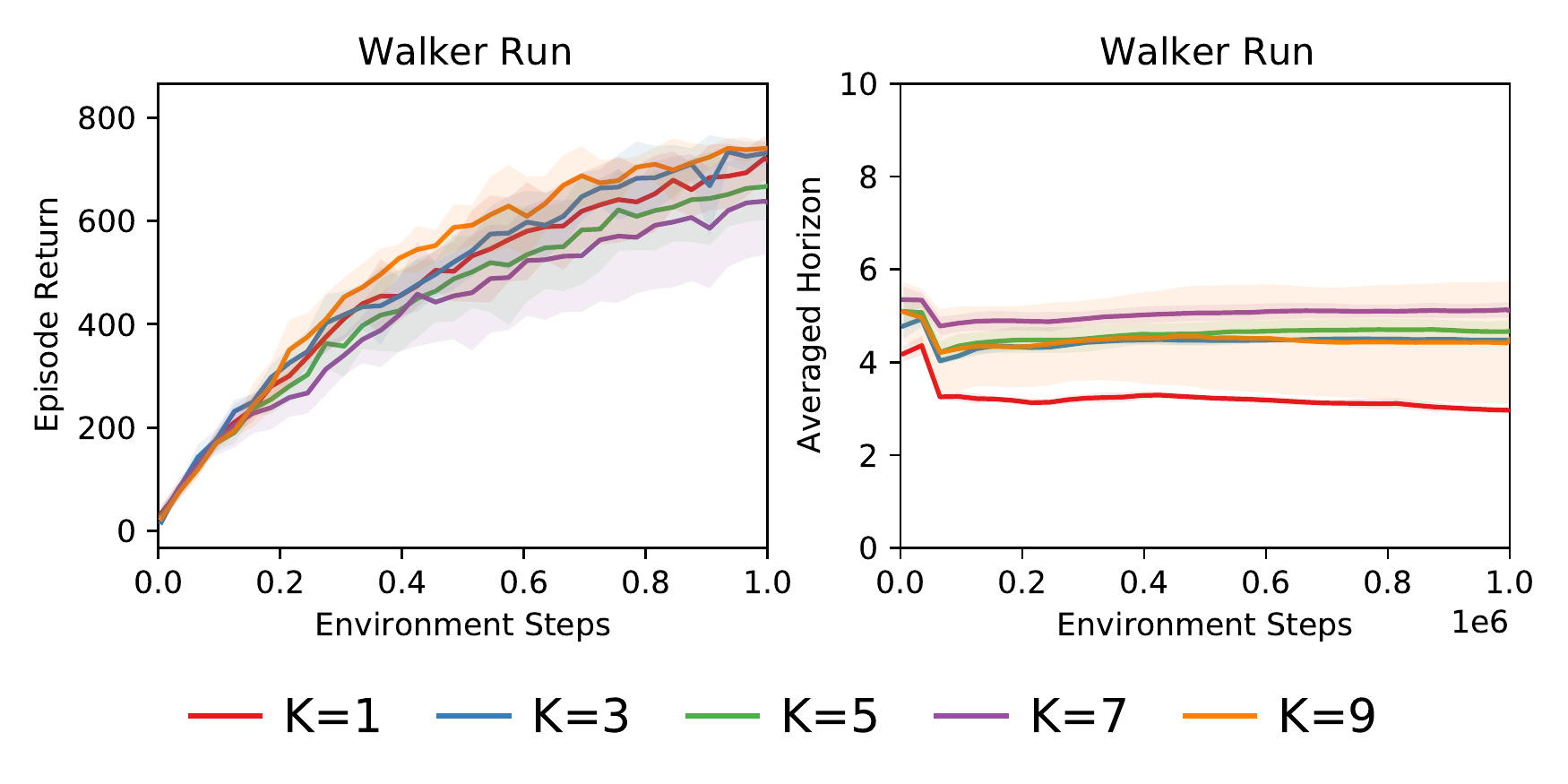}}
\caption{The learning curves and the averaged horizon curves of DMVE with different number of selected horizons $K$ on the task of Walker Run. The solid lines depict the mean scores and shaded areas indicate the standard deviation of five trials over different random seeds. 
Each trial is evaluated every 10,000 environment steps. 
}
\label{DMVEK}
\end{figure}

In addition, we vary $K$ to investigate the sensitivity of DMVE to it ($H$ is fixed to $15$ across different runs). The results are plotted in Figure~\ref{DMVEK}. We find that different $K$-value choices have some effect on the results, but overall DMVE can learn a good policy on the Walker Run task. Furthermore, we also find that although the number of selected horizons is diverse, the average value of horizons selected varies relatively little for different $K$, which reflects the stability of horizon selection with reconstruction-based methods. Moreover, the selection of multiple horizons also collects different horizon values into final value estimation like $V_ \lambda$ does, but horizon selection works well with DMVE, not $V_ \lambda$. One possible reason is that $V_ \lambda$ incorporates long-horizon information even the maximum, which may harm the value estimation due to the model-bias. On the contrast, the horizons selected by DMVE are relatively small even with relatively large $K$ (e.g. $K=9$). 

% \begin{figure}[ht]
% \subfigure[The test return of individual episodes.]{
% \includegraphics[width=3.0in]{DMVEK}
% }
% \subfigure[The mean value of selected horizons.]{
% \includegraphics[width=3.0in]{HMean}
% }
% \caption{The learning curves of DMVE with different number of selected horizons $K$ on the task of Walker Run. The solid lines depict the mean scores and shaded areas indicate the standard deviation of t3hree trials over different random seeds. Each trial is evaluated every 10,000 environment steps. }
% \label{DMVEK}
% \end{figure}

% \begin{figure}[ht]
% \centering
% \includegraphics[width=0.9\columnwidth]{curves} % Reduce the figure size so that it is slightly narrower than the column. Don't use precise values for figure width.This setup will avoid overfull boxes.
% \caption{Using the trim and clip commands produces fragile layers that can result in disasters (like this one from an actual paper) when the color space is corrected or the PDF combined with others for the final proceedings. Crop your figures properly in a graphics program -- not in LaTeX}.
% \label{curves}
% \end{figure}

\section{Conclusion}
In this paper, we present an advanced method Dynamic-horizon Model-based Value Expansion (DMVE) which can adjust the world model usage for value estimation with adaptive state-related imagination horizons. We develop an algorithm which incorporates model-based value expansion with latent imagination. Also, an adaptive rollout horizon selection method is designed based on it. Experimental results indicate that DMVE outperforms state-of-the-art model-based method Dreamer on several benchmark visual control tasks. Since DMVE leverages reconstruction-based models that are limited to pixel inputs, our future work includes investigating the dynamic world model usage for state input tasks.

\bibliography{bib}

\end{document}